\def\year{2021}\relax
\documentclass[letterpaper]{article} % DO NOT CHANGE THIS
\usepackage{aaai21}  % DO NOT CHANGE THIS
\usepackage{times}  % DO NOT CHANGE THIS
\usepackage{helvet} % DO NOT CHANGE THIS
\usepackage{courier}  % DO NOT CHANGE THIS
\usepackage[hyphens]{url}  % DO NOT CHANGE THIS
\usepackage{graphicx} % DO NOT CHANGE THIS
\usepackage{natbib}  % DO NOT CHANGE THIS AND DO NOT ADD ANY OPTIONS TO IT
\usepackage{caption} % DO NOT CHANGE THIS AND DO NOT ADD ANY OPTIONS TO IT
\usepackage{color}
\usepackage{cite}
\usepackage{amsmath}
\usepackage{amssymb}
\usepackage{subfigure}
\usepackage{color}
\usepackage{epsfig}
\usepackage{epstopdf}
\usepackage{algorithm}
\usepackage{algorithmic}
\usepackage{multirow}
\usepackage{comment}
\usepackage{diagbox}
\usepackage{hyperref} 

\title{Learning Geometry-Disentangled Representation for Complementary Understanding of 3D Object Point Cloud}
\author{
    Mutian Xu \textsuperscript{\rm 1 2}\thanks{M.Xu and J.Zhang contribute equally.}, 
	Junhao Zhang \textsuperscript{\rm 1}\footnotemark[1],
	Zhipeng Zhou \textsuperscript{\rm 1},
	Mingye Xu \textsuperscript{\rm 1 3},\\
	Xiaojuan Qi \textsuperscript{\rm 2},
	Yu Qiao \textsuperscript{\rm 1}\thanks{Corresponding author.}\\
}
\affiliations{
    \textsuperscript{\rm 1}Guangdong-Hong Kong-Macao Joint Laboratory of Human-Machine Intelligence-Synergy Systems,\\ Shenzhen Institutes of Advanced Technology, Chinese Academy of Sciences\\%，Shenzhen, 518055, China\\
    \textsuperscript{\rm 2}The University of Hong Kong\\
    \textsuperscript{\rm 3}University of Chinese Academy of Sciences\\
    mino1018@outlook.com, \{zhangjh, zp.zhou, my.xu\}@siat.ac.cn, xjqi@eee.hku.hk, yu.qiao@siat.ac.cn
}

\begin{document}
%\linenumbers 

\maketitle

\begin{abstract}
In 2D image processing, some attempts decompose images into high and low frequency components for describing edge and smooth parts respectively. Similarly, the contour and flat area of 3D objects, such as the boundary and seat area of a chair, describe different but also complementary geometries. However, such investigation is lost in previous deep networks that understand point clouds by directly treating all points or local patches equally. To solve this problem, we propose Geometry-Disentangled Attention Network (GDANet). GDANet introduces Geometry-Disentangle Module to dynamically disentangle point clouds into the contour and flat part of 3D objects, respectively denoted by sharp and gentle variation components. Then GDANet exploits Sharp-Gentle Complementary Attention Module that regards the features from sharp and gentle variation components as two holistic representations, and pays different attentions to them while fusing them respectively with original point cloud features. In this way, our method captures and refines the holistic and complementary 3D geometric semantics from two distinct disentangled components to supplement the local information. Extensive experiments on 3D object classification and segmentation benchmarks demonstrate that GDANet achieves the state-of-the-arts with fewer parameters. Code is released on \href{https://github.com/mutianxu/GDANet}{https://github.com/mutianxu/GDANet}.
\end{abstract}

\section{Introduction}
The capacity to analyze and comprehend 3D point clouds receives interests in computer vision community due to its wide applications in autonomous driving and robotics \cite{rusu2008towards,qi2018frustum}. Recent studies explore deep learning methods to understand 3D point clouds inspired by their great success in computer vision applications \cite{He_2015_ICCV,He_2016_CVPR}. Deep networks \cite{7569026} can extract effective semantics of 3D point clouds with layered operations, in contrast to low-level handcrafted shape descriptors. 
\begin{figure}[t]
	\setlength{\abovecaptionskip}{0cm} 
	\begin{center}
	\includegraphics[height=2.35cm]{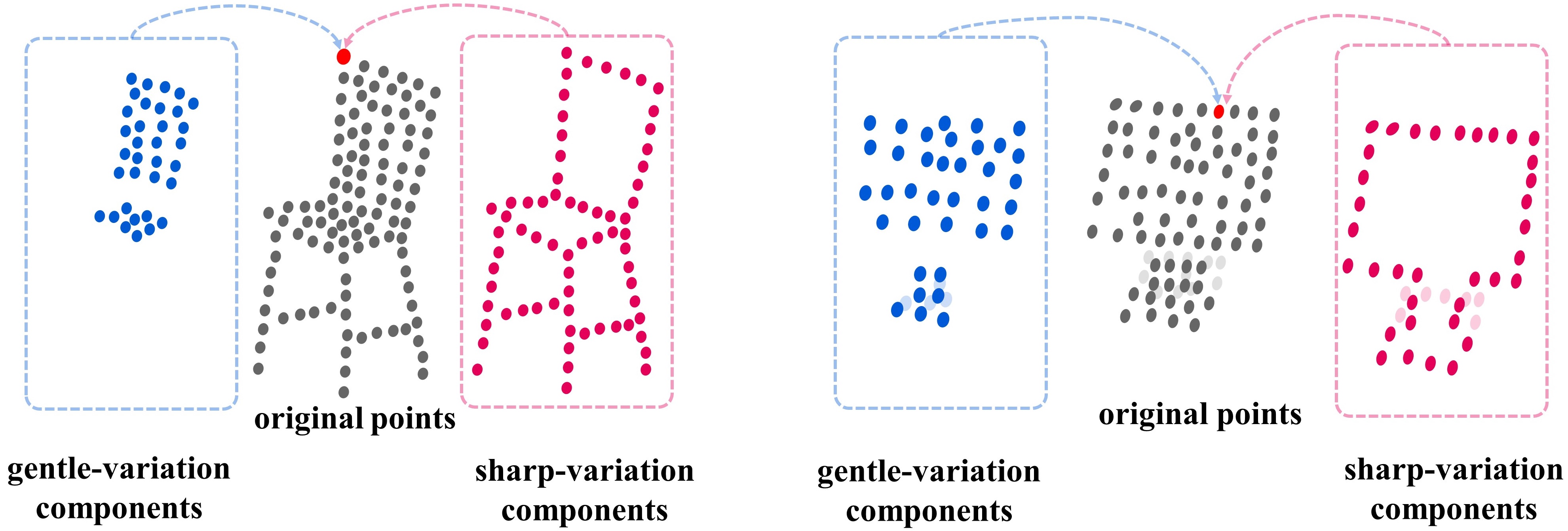} % Reduce the figure size so that it is slightly narrower than the column. Don't use precise values for figure width.This setup will avoid overfull boxes.
	\end{center}
	\caption{Examples of 3D objects, where sharp-variation component describes the contour areas, and gentle-variation component denotes the flat areas. Our method regards the features from these two disentangled components as two holistic representations, i.e., each point of the original point cloud is linked with all points of sharp and gentle variation components. This operation integrates complementary geometric information from two disentangled components.}
	\label{first}
\end{figure}
The pioneer work PointNet \cite{Qi_2017_CVPR} directly processes 3D points by Multi-layer Perceptrons(MLPs) \cite{MLP}, whose main idea is to learn a spatial encoding of each point and then aggregate all point features to a global point cloud signature. %However, it lacks the ability to capture local structures of 3D manifolds. Several works try to address this problem.
PointNet++ \cite{NIPS2017_7095} adopts a hierarchical encoder-decoder structure to consider local regions, which downsamples point clouds in layers and gradually interpolates them to the original resolution. From another perspective, some recent efforts extend regular grid convolution \cite{Xu_2018_ECCV,NIPS2018_7362,thomas2019KPConv} on irregular 3D point cloud configuration. 

%The mentioned methods largely neglect the importance of geometric structure in modeling point clouds. To remedy this defect, 
To fully utilize geometric information, some attempts \cite{Wang:2019:DGC:3341165.3326362,Lan_2019_CVPR} capture local geometric relations among center points and its neighbors. However, these works treat all points or local patches equally, which are entangled together with large redundancy, making it hard to capture the most related and key geometric interest to the network. Moreover, previous operations only capture the geometric information in local areas.

To remedy these defects, we need to disentangle point clouds into distinct components and learn the few-redundant information represented by these holistic components. In image processing, some attempts collect and combine the high-frequency (edge) and low-frequency (smooth) components with distinct characteristics filtered through digital signal processing. Similarly, the contour areas of 3D objects delineating skeletons provide basic structural information, while the flat areas depicting surfaces supply the geometric manifold context. When operating them separately, the network performs better by learning distinct but also complementary geometric representations of point clouds. This inspires us to extract and utilize the geometric information of the contour and flat areas disentangled from 3D objects.
%the contour and flat areas of 3D objects play distinct and complementary roles in the geometric representation of point clouds, which inspires us to extract and utilize the distinct geometric information of the contour and flat areas disentangled from 3D objects.

Here comes a challenge about how to disentangle 3D objects into such holistic representations (contour and flat area) and utilize them for better understanding point clouds. Thanks to the graph signal processing \cite{graphsignal,discrete} who analyzes the frequency on graphs, we firstly extend this to our Geometry-Disentangle Module (GDM) that dynamically analyzes graph signals on 3D point clouds in different semantic levels and factorizes the original point cloud into the contour and flat parts of objects, respectively denoted by sharp and gentle variation components (explained at the end of Sec. 4.1). Further, we design Sharp-Gentle Complementary Attention Module (SGCAM) that pays different attentions to features from sharp and gentle variation components according to geometric correlation, then respectively fuses them with each original point features instead of only operating local patches. As shown in Fig.~\ref{first}, the contour and flat area have distinct but also complementary effects on reflecting the geometry of objects. 

Equipped with GDM and SGCAM, we propose GDANet who captures and refines the holistic and complementary geometries of 3D objects to supplement local neighboring information. Experimental results on challenging benchmarks demonstrate that our GDANet achieves the state of the arts, and is more lightweight and robust to density, rotation and noise. Thorough visualizations and analysis verify that our model effectively captures the complementary geometric information from two disentangled components.

%We summarize the key contributions of this work as follows:

 %$\bullet$ We propose Geometric Selection Module that selects points with large and small geometric variation (respectively refer to the most representative points and relatively common points) of a point cloud, which can be easily incorporated into existing pipelines for point cloud analysis.  
 
% $\bullet$ We present LSCAM that applies the attention mechanism to fuse the complementary features from large and small geometric variation points with original point features. Equipped with GSM and LSCAM, we design GDANet who captures and refines the holistic and complementary geometries of 3D objects to supplement local neighboring information.
 
 %$\bullet$ Experimental results on challenging benchmarks demonstrate that our GDANet achieves the state of the arts, and is more lightweight and robust to density, rotation and noise. Thorough visualizations and analysis also verify that our model effectively captures the complementary geometric information.

\section{Related Work}
%\noindent{\textbf{Deep Learning on Point Clouds.}} 
%\subsubsection{Deep Learning on Point Clouds.}
%PointNet \cite{Qi_2017_CVPR} is the pioneer who processes the 3d point clouds through learning each individual point equally. Nonetheless, it neglects the importance of modeling local patches. Thus, some methods \cite{NIPS2017_7095,pointweb,Shen_2018_CVPRkc} aggregate point features in local areas for better representing regional structures. Additionally, to solve the irregularity issue of points, \cite{PointConv,thomas2019KPConv,NIPS2018_7362,interconv,fpconv} focuses on applying regular CNN on irregular 3D point cloud configuration. Convolution operator can also be applied on graph with features from spatial information directly \cite{Wang_2018_ECCVspec,Wang:2019:DGC:3341165.3326362}. Point features and edge features are aggregated through graph convolution, which focuses on capturing local surface information while being invariant to the transformation of those patches. However, these methods fail to explicitly model the 3d geometric structure, which our humans largely rely on in 3D shape-aware understanding. Instead, our network extracts two different geometric parts that provide complementary geometric semantics for understanding 3d objects. 

\noindent{\textbf{Point Cloud Models Based on Geometry.}} 
%\subsubsection{Point Cloud Models Based on Geometric Information.}
Recently the exploration on point cloud geometry has drawn large focus. Geo-CNN \cite{Lan_2019_CVPR} defines a convolution operation that aggregates edge features to capture the local geometric relations. In \cite{xu2020geometry}, they aggregate points from local neighbors with similar semantics in both Euclidean and Eigenvalue space. RS-CNN \cite{liu2019rscnn} extends regular CNN to encode geometric relation in a local point set by learning the topology. DensePoint \cite{liu2019densepoint} recognizes the implicit geometry by extracting contextual shape semantics. However, the contour and flat area of 3D objects play different but complementary roles in modeling 3D objects. All the operations mentioned above neglect this property and treat all points or local patches equally. By contrast, we present Geometry-Disentangle Module to disentangle point clouds into sharp (contour) and gentle (flat area) variation components, which are two distinct representations of 3D objects. 
% Then our Large-Small Complementary Attention Module pays different attentions to these two components while connecting each point in the original point cloud respectively with all points from these two components instead of only operating on local areas. 

\noindent{\textbf{Attention Networks.}}
The applications of attention mechanism in sequence-based tasks become popular \cite{NIPS2017_7181}, which helps to concentrate on the most relevant and significant parts. Some recent point cloud methods utilize attention to aggregate the neighboring features of each point \cite{point2sequence}. GAC \cite{gac_2019_CVPR} proposes a graph attention convolution that can be carved into specific shapes by assigning attention weights to different neighbor points. 
%Different from them, our aim of using attention is to fuse the original point cloud features with features from sharp and gentle variation components respectively. This provides us an effective fusion approach of complementary geometric information by assigning different weights to relevant points based on geometric correlation.
Different from them, our network learns to assign different attention weights to the disentangled contour and flat areas of 3D objects based on geometric correlations. In Sec. 4.3, we also illustrate that not only the attention mechanism helps the network refine the disentangled feature but also our disentanglement strategy assists the attention module easily concentrate on the key geometric interests.

\noindent{\textbf{Disentangled Representation.}}
Recent attempts use disentangled representations in different applications. In general, the concept of disentanglement \cite{bengio2012representation} dominates representation learning, closely linking with human reasoning. In \cite{Xiao_2018_ECCV}, they separate a facial image into individual and shared factors encoding single attribute. \cite{Huang_2018_ECCV} processes images by decomposing latent space into content and style space. 
%Although these works utilize disentangled representations to decompose data into different informative elements, 
Yet, the disentangled representation on point cloud understanding remains untouched. Our method explicitly disentangles point clouds into two components denoting contour and flat area of objects, which are fused to provide distinct and complementary geometric information. % To the best of our knowledge, we are the first to learn the explicit geometry-disentangled representation for understanding point clouds.

\section{Revisit Graph Signal Processing}
 Graph signal processing \cite{graphsignal,discrete} is based on a graph $\mathcal{G}=(\mathcal{V},\bf A)$ where $\mathcal{V}=\{v_1,\cdots,v_{N}\}$ denotes a set of $N$ nodes and $\mathbf A \in \mathbb{R}^{N\times N}$ denotes a weight adjacency matrix encoding the dependencies between nodes. Using this graph, we refer to the one-channel features of the data on all nodes in vertex domain as $\mathbf{s} \in \mathbb{R}^{N}$.
Let $\mathbf{A}$ be a graph shift operator which takes a graph signal $\mathbf{s} \in \mathbb{R}^N$ as input and produces a new graph signal $\mathbf {y}= \mathbf{A} \mathbf{s}$.
We also have the eigen decomposition 
$\mathbf {A}={V} \mathbf{\Lambda} {V}^{-1}$, where the columns of matrix ${V}$ are the eigenvectors of $\bf A$ and the diagonal eigenvalue matrix $\mathbf{\Lambda} \in \mathbb{R}^{N \times N} $ corresponds to ordered eigenvalues $\lambda_1,\cdots,\lambda_{N}$. 

 \vspace*{0.4\baselineskip}
\noindent\textbf{Theorem 1} \cite{graphsignal,discrete}.
$The$ $ordered$ $eigenvalues$ $(\lambda_{1} \geq \lambda_{2} \geq \cdots, \geq \lambda_N)$ $represent$ $frequencies$ $on$ $the$ $graph$ $from$ $low$ $to$ $high$.

 \vspace*{0.4\baselineskip}
Accordingly, we obtain ${V}^{-1}\mathbf{y} = \mathbf{\Lambda} {V}^{-1} \bf s$, and
     the graph Fourier transform of graph signal $\bf s$ and $\bf y$ are $\widehat{\mathbf{s}}=V^{-1} \mathbf{s}$, $\widehat{\mathbf{y}}=V^{-1} \mathbf{y}$, respectively. $V^{-1}$ is the graph Fourier transform matrix. The components of
      $\widehat{\mathbf{s}}$ and $\widehat{\mathbf{y}}$ are considered as frequency contents of signal $\bf s$ and $\bf y$.
      
As stated in \cite{discrete}, a graph filter is a polynomial in the graph shift: \begin{equation}
h(\mathbf{A})=\sum_{\ell=0}^{L-1} h_{\ell} \mathbf{A}^{\ell},
\label{111}
\end{equation}
where $h_{\ell}$ are filter coefficients and $L$ is the length of the filter. It takes a graph signal $\mathbf s \in \mathbb{R}^{N} $ as the input and generates a filtered signal $\mathbf{y}=h(\mathbf A)s \in \mathbb{R}^{N}$. Then $\mathbf {y}={V} {h}(\mathbf{\Lambda}){V}^{-1}\bf s$, making $V^{-1}\mathbf {y}= {h}(\mathbf{\Lambda}) {V}^{-1}\bf s$ and $\widehat{\mathbf{y}}={h}(\mathbf{\Lambda})\widehat{\mathbf{s}}$. 

\vspace*{0.4\baselineskip}
\noindent\textbf{Theorem 2} \cite{graphsignal,discrete}. $The$ $diagonal$ $matrix$ $h(\mathbf{\Lambda})$ $is$ $the$ $graph$ $frequency$ $response$ $of$ $the$ $filter$ $h(\bf{A}),$ $which$ $is$ $denoted$ $as$ $\widehat{h(\bf{A})}$. $The$ $frequency$ $response$ $of$ $\lambda_i$ $is \sum_{\ell=0}^{L-1} h_{\ell} \lambda_{i}^{\ell}$.
\section{Method}
We firstly design Geometry-Disentangle Module based on graph signal processing to decompose point clouds into sharp and gentle variation components (respectively denotes the contour and flat area). Further, we propose Sharp-Gentle Complementary Attention Module to fuse the point features from sharp and gentle variation components. Last, we introduce Geometry-Disentangled Attention Network equipped with these two modules.

\subsection{4.1 Geometry-Disentangle Module}\label{4.1}

\subsubsection{Graph Construction.}

We consider a point cloud consisted of $N$ points with $C-$dimensional features, which is denoted by a matrix $X=[x_{1}, x_{2}, \cdots  ,x_{N}]^{T}=[\mathbf{s}_{1}, \mathbf{s}_{2}, \cdots,\mathbf{s}_{C}]\in \mathbb{R}^{N \times C}$, where $x_i \in  \mathbb{R}^{C}$ represents the $i$-th point and $\mathbf{s}_{c} \in \mathbb{R}^{N}$ represents the $c$-th channel feature. Features can be 3D coordinates, normals and semantic features. We construct a graph $\mathcal{G}=(\mathcal{V},\mathbf{A})$ through an adjacency matrix $\mathbf{A}$ that encodes point similarity in the feature space. Each point $x_{i} \in \mathbb{R}^{C}$ is associated with a corresponding graph vertex $i \in \mathcal{V}$ and $\mathbf{s}_{c} \in \mathbb{R}^{N}$  is a graph signal. The edge weight between two points $x_i$ and $x_j$ is
\begin{equation}
\mathbf{A}_{i,j}=
\begin{cases}
f(||x_i-x_j||_{2}),& \text{$||x_i-x_j||_2 \leq \tau$}\\
0,& \text{otherwise}
\end{cases},
\end{equation}
where $f(\cdot)$ is an non-negative decreasing function (e.g., Gaussian function) which must ensure that $\textbf{A}\in \mathbb{R}^{N \times N}$ is a diagonally dominant matrix and $\tau$ is a threshold. In addition, to handle the size-varying neighbors across
different points and feature scales, we normalize all edge weights as follows:
\begin{equation}
 \mathbf{\tilde{A}}_{i ,j}=\frac{ \mathbf{A}_{i,j}}{\sum_{j}  \mathbf{A}_{i,j}},
\end{equation}
where $\mathbf{\tilde{A}}$ is still a diagonally dominant matrix. Now as illustrated in Theorem 1, we obtain a graph  $\mathcal{G}=(\mathcal{V},\mathbf{\tilde{A}})$, where
eigenvalues of  $\mathbf{\tilde{A}}$ $\left(\tilde{\lambda}_{1} \geq \tilde{\lambda}_{2} \geq \cdots, \geq \tilde{\lambda}_N \right)$ represent graph frequencies from low to high.

%\noindent{{\bf Selection of large and small geometric variation points.}} 

\subsubsection{Disentangling Point Clouds into Sharp and Gentle Variation Components.}
In 2D image processing, high frequency component corresponding to intense pixel variation (edge) in spatial domain gets high response while low frequency component (smooth area) gets very low response after being processed by a high-pass filter. Here we aim to design a graph filter on our constructed $\mathcal{G}=(\mathcal{V},\mathbf{\tilde{A}})$ to select the points belongs to the contour and flat areas of 3D objects. 
\begin{figure}[ht]
\setlength{\abovecaptionskip}{0cm} 
	\begin{center}
	\includegraphics[height=3.62cm]{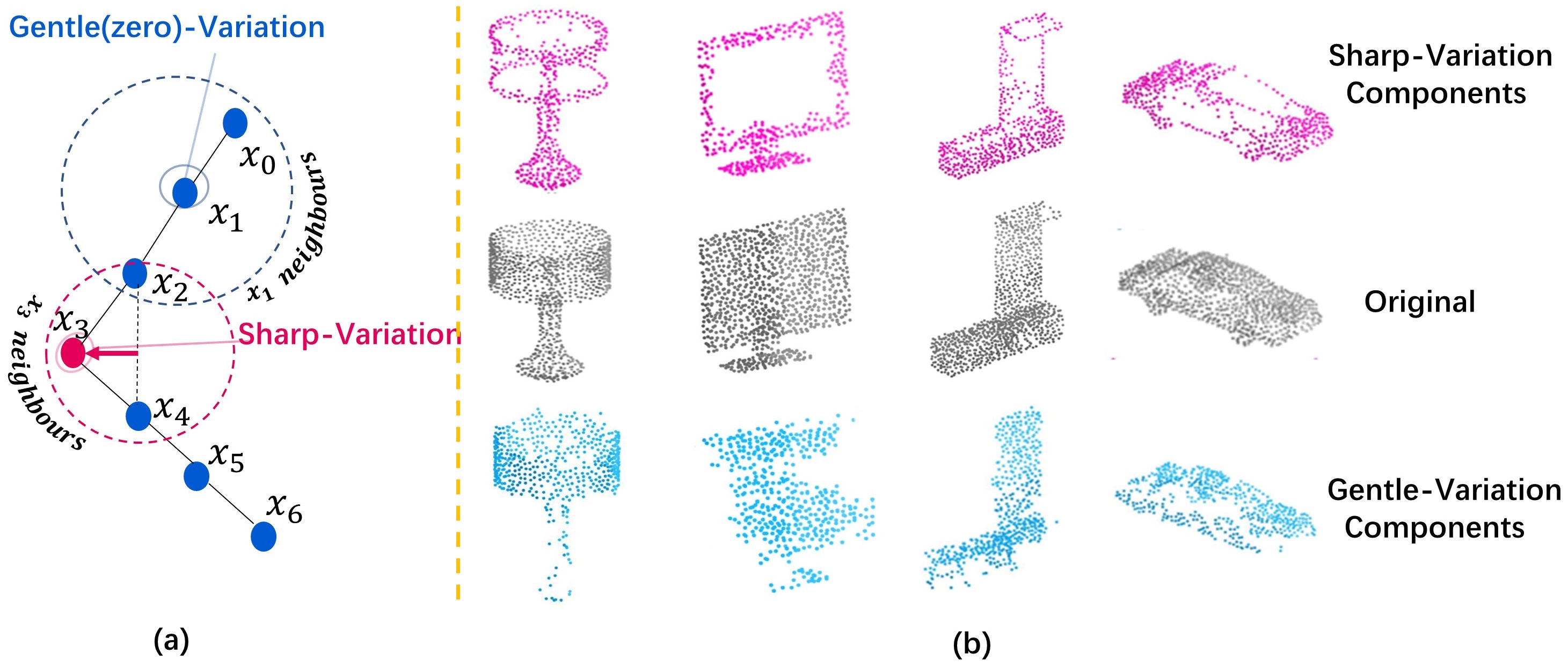} % Reduce the figure size so that it is slightly narrower than the column. Don't use precise values for figure width.This setup will avoid overfull boxes.
	\end{center}
	\setlength{\belowcaptionskip}{0cm} 
	\caption{Visualization of the process in our Geometry-Disentangle Module and the obtained sharp and gentle variation components of some objects.}
	\label{selection}
\end{figure}
Following Eq.~(\ref{111}), the key to design this graph filter is to construct the corresponding polynomial format of $h(\mathbf{\tilde{A}})$.
%$h(\mathbf{\tilde{A}})=\sum_{\ell=0}^{L-1} h_{\ell} \mathbf{\tilde{A}}^{\ell}$. 
Here we use the Laplacian operator as \cite{fast}, where $L=2$, $h_0=1$, $h_1=-1$, making the polynomial format of the graph filter to be $h(\mathbf{\tilde{A}})=I-\mathbf{\tilde{A}}$. This filter takes $\mathbf{s}_{c} \in \mathbb{R}^N$ in this graph as the input and generates a filtered graph signal $\mathbf {y}_{c}= h(\mathbf{\tilde{A}}) \mathbf{s}_{c} \in \mathbb{R}^N$. Following Theorem 2, the frequency response of $h(\mathbf{\tilde{A}})$ with corresponding $\lambda_{i}$ is
\begin{equation}
\widehat{h(\bf \tilde{A})}=\begin{tiny}\left[\begin{array}{cccc}{1-\tilde{\lambda}_{1}} & {0} & {\cdots} & {0} \\ {0} & {1-\tilde{\lambda}_{2}} & {\cdots} & {0} \\ {\vdots} & {\vdots} & {\ddots} & {\vdots} \\ {0} & {0} & {\cdots} & {1-\tilde{\lambda}_{N}} \end{array}\right]
\end{tiny}.\end{equation} 
In this way, the eigenvalues $\tilde{\lambda}_{i}$ are in a descending order, which represents that frequencies are ordered ascendingly according to Theorem 1. Due to the corresponding \textbf{frequencies response} $1-\tilde{\lambda}_{i} < 1-\tilde{\lambda}_{i+1}$, the low frequency part is weakened after this filter. Hence, we call this filter $h(\mathbf{\tilde{A}})=I-\mathbf{\tilde{A}}$ a \textbf{high-pass} filter. 

\textit{\textbf{Note}}: The eigenvalues representing frequencies is only for deducing the polynomial format of our high-pass filter $h(\mathbf{\tilde{A}})$. The implementation of this filter only requires the calculation of $\mathbf{\tilde{A}}$.

Next, we apply $h(\mathbf{\tilde{A}})$ to filter the point set $X$ and get a filtered point set $h(\mathbf{\tilde{A}})X$. Due to $h(\mathbf{\tilde{A}})=I-\mathbf{\tilde{A}}$, each point in $h(\mathbf{\tilde{A}})X$ can be formulated as: 
\begin{equation} 
    \left(h(\mathbf{\tilde{A}}) \mathrm{X}\right)_{i}=\mathrm{x}_{i}-\sum_{j}^{N} \mathbf{\tilde{A}}_{i, j} \mathrm{x}_{j}.
    \label{final}
\end{equation}
When the distance between two point $x_{i},x_{j}$ is less than the threshold $\tau$, $\mathbf{A}_{i,j}$ remains non-zero value. Here $\left(h(\mathbf{\tilde{A}}) \mathrm{X}\right)_{i}$ actually equals to the difference between a point feature and the linear convex combination of its neighbors' features, which reflects the degree of each point's \textbf{variation} to its neighbors.

Finally, we calculate the $l^2$-norm of Eq. (\ref{final}) at every point, and the larger $l^2$-norm at a given point reflects sharp variation and means this point belongs to the \textbf{contour} of a 3D object, which is consistent to the edge areas obtained by a high-pass filter in 2D images. We put all original points in descending order as $X_o=[\check{x}_{1}, \check{x}_{2}, \cdots  ,\check{x}_{N}]^{T}$ according to $l^2$-norm of Eq. (\ref{final}). Following this order, we select the first $M$ points $X_s=[\check{x}_{1}, \check{x}_{2}, \cdots, \check{x}_{M}]^{T} \in \mathbb{R}^{M \times C}$ called as \textbf{sharp-variation component} and the last $M$ points \(X_g=[\check{x}_{N-M+1}, \check{x}_{N-M+2},\cdots, \check{x}_{N}]^{T} \in \mathbb{R}^{M \times C}\) denoted by \textbf{gentle-variation component}. Fig.~\ref{selection} (a) shows this process and Fig.~\ref{selection} (b) visualizes sharp and
gentle variation components disentangled by our trained network. We employ our Geometry-Disentangle Module on point features in different semantic levels, which is elaborated in Sec. 4.4.

\begin{figure}[t]
\setlength{\abovecaptionskip}{0cm} 
	\begin{center}
	\includegraphics[height=4.0cm]{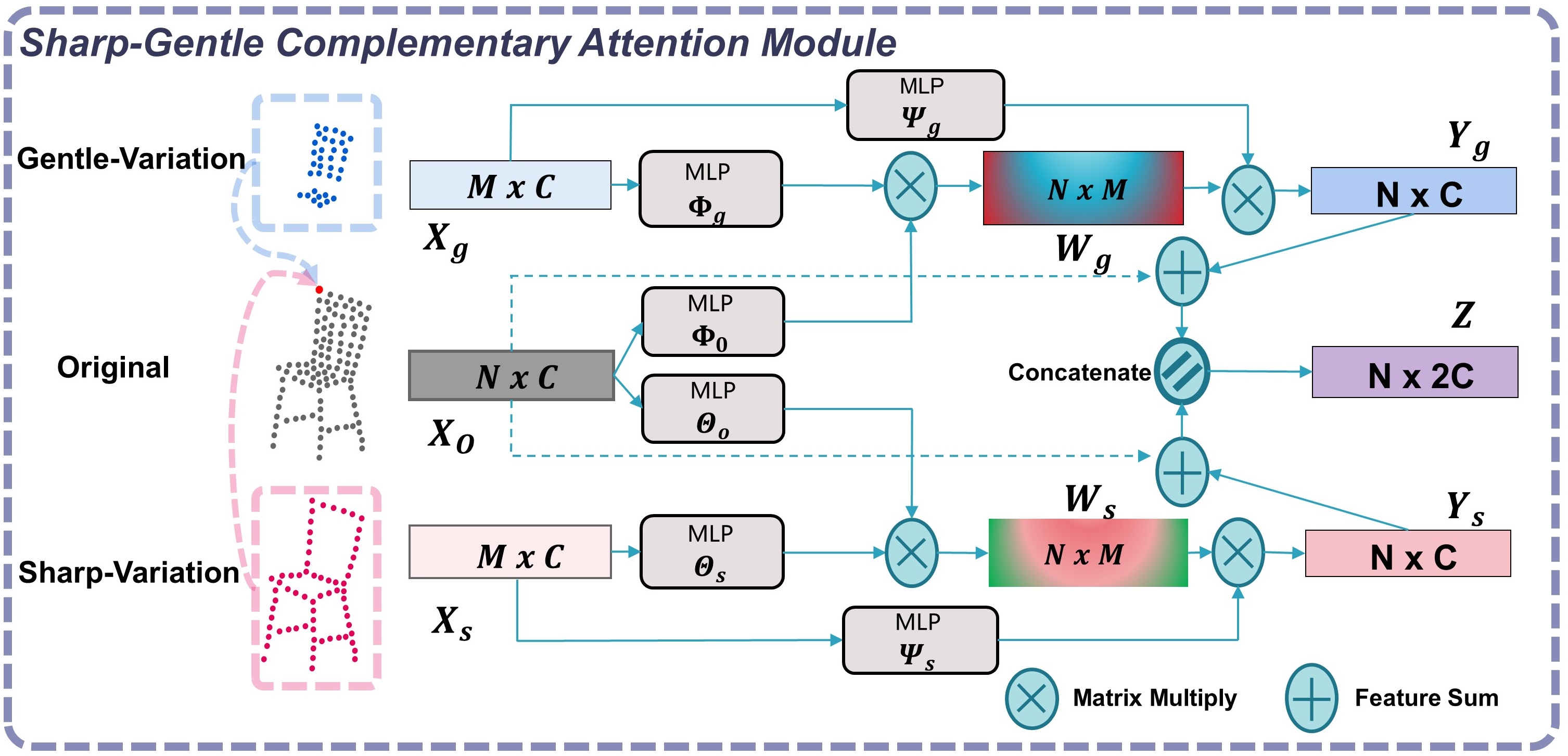} % Reduce the figure size so that it is slightly narrower than the column. Don't use precise values for figure width.This setup will avoid overfull boxes.
	\end{center}
	\caption{Visualization of Sharp-Gentle Complementary Attention Module. Features from different variation components are respectively integrated with the features from original point cloud, so that to provide complementary geometric information.}
	\label{LSmodule}
\end{figure}

\subsection{4.2 Sharp-Gentle \hspace{-0.4mm}Complementary\hspace{-0.5mm} Attention \hspace{-0.5mm}Module}
The sharp and gentle variation components play different but complementary roles in representing the 3D object geometries, which should not be treated equally. However, most previous methods operate all points or local point sets equally. To solve this issue and utilize different variation components gained from Geometry-Disentangle Module, we design Sharp-Gentle Complementary Attention Module inspired by \cite{NIPS2017_7181,nonlocal}. It regards the features from two variation components as two holistic geometric representations and generates two attention matrices separately according to the geometric correlation. Then our module assigns the corresponding attention weights to features from two different variation components while respectively integrating them with the original input point features. The details of Sharp-Gentle Complementary Attention Module are shown in Fig.~\ref{LSmodule} and elaborated below. 

\subsubsection{Attention Matrix.}
According to Sec. 4.1, we have original point cloud features $X_{o}$, features of sharp-variation component $X_{s}$ and features of gentle-variation component $X_{g}$. 
These features are firstly encoded by different nonlinear functions, and then are utilized to calculate different attention matrices as the following equation:
\begin{equation}
%	W^{l}=f_{o}^{1}(\tilde{X}) \cdot f_{l}^{1}\left(X_{l}\right)^{T};	W^{s}=f_{o}^{2}(\tilde{X}) \cdot f_{s}^1\left(X_{s}\right)^{T}.
W_{s}=\Theta_{o}(X_{o}) \cdot \Theta_{s}\left(X_{s}\right)^{T},	W_{g}=\Phi_{o}(X_{o}) \cdot \Phi_{g}\left(X_{g}\right)^{T},
	\label{weight_h}
\end{equation}
where different nonlinear functions $\Theta_{o}$, $\Theta_{s}$, $\Phi_{o}$ and $\Phi_{g}$ are implemented by different MLPs without sharing parameters. In this way, we get two learnable adjacency matrices \(W_{s} \in \mathbb{R}^{N\times M} \) and \(W_{g} \in \mathbb{R}^{N\times M} \), where $M$ is the number of points in either \(X_{s}\) or \(X_{g}\). Each row of $W_{s}$ or $W_{g}$ corresponds to attention weights between each original point feature and features from sharp and gentle variation components, respectively. Because the adjacency matrices \(W_{s}\) and \(W_{g}\) are computed as feature dot product, they can explicitly measure the semantic correlation or discrepancy between points.

\begin{figure*}[t]
	\begin{center}
	\includegraphics[width=1.9\columnwidth]{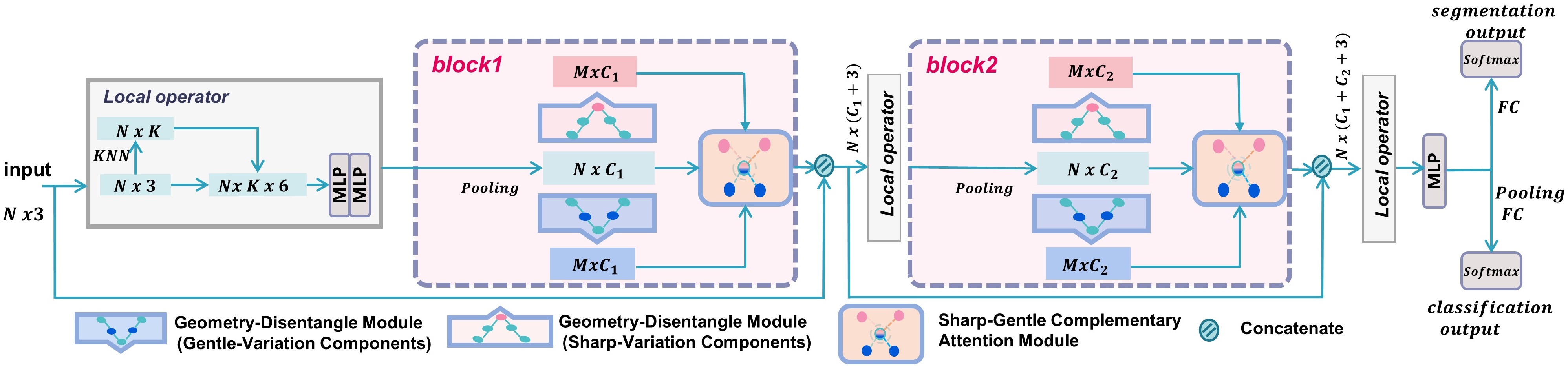}
	\end{center}
	
    \setlength{\belowcaptionskip}{0.cm}
	\caption{GDANet architecture for classification and segmentation. Our network disentangles the original point cloud into sharp and gentle variation components in different semantic levels, then fuses features from these two components with the input point features to supplement the KNN local context.}
	\label{FAFNET}
\end{figure*}

%\noindent{\bf Geometric complementary representation.} 
\subsubsection{Geometric Complementary Understanding.}
Next we apply the attention matrices $W_{s}$ and $W_{g}$ respectively to the features from sharp and gentle variation components so that the network can pay different attentions to them while fusing them with the original point features. The whole fusion procedure can be formulated as the following:
\begin{equation}
	Y_{s} = X_{o} + W_{s} \cdot \Psi_{s}\left(X_{s}\right), 
	\label{y_h}
\end{equation}
\begin{equation}
	Y_{g} = X_{o} + W_{g} \cdot \Psi_{g}\left(X_{g}\right), 
	\label{y_l}
\end{equation}
in element-wise:
\begin{equation}
	(Y_{s})_{i}=(X_{o})_{i}+\sum\nolimits_{j=1}^{M} \left(W_{s}\right)_{i j} \cdot \Psi_{s}((X_{s})_{j}),
	\label{y_h_e}
\end{equation}
\begin{equation}
	(Y_{g})_{i}=(X_{o})_{i}+\sum\nolimits_{j=1}^{M} (W_{g})_{i j} \cdot \Psi_{g}((X_{g})_{j}),\hspace{-3.8807mm}
	\label{y_h_e}
\end{equation}
where two different nonlinear functions $\Psi_{s}$ and $\Psi_{g}$ are utilized to refine $X_{s}$ and $X_{g}$. They are implemented by different MLPs without sharing parameters.

Last we concatenate the features as the following equation:
\begin{equation}
	Z=Y_{s} \oplus Y_{g}.
\end{equation}

Accordingly, our method regards features from sharp and gentle variation components as two holistic representations, i.e., all the point features with different attention weights from these two components are respectively linked with each original input point cloud feature. Our module explicitly conveys the complementary geometric information and the most relevant and key geometric interest to the network in a holistic way for better understanding 3D point cloud.
\begin{figure}[t]
	\begin{center}
	\includegraphics[width=1\columnwidth]{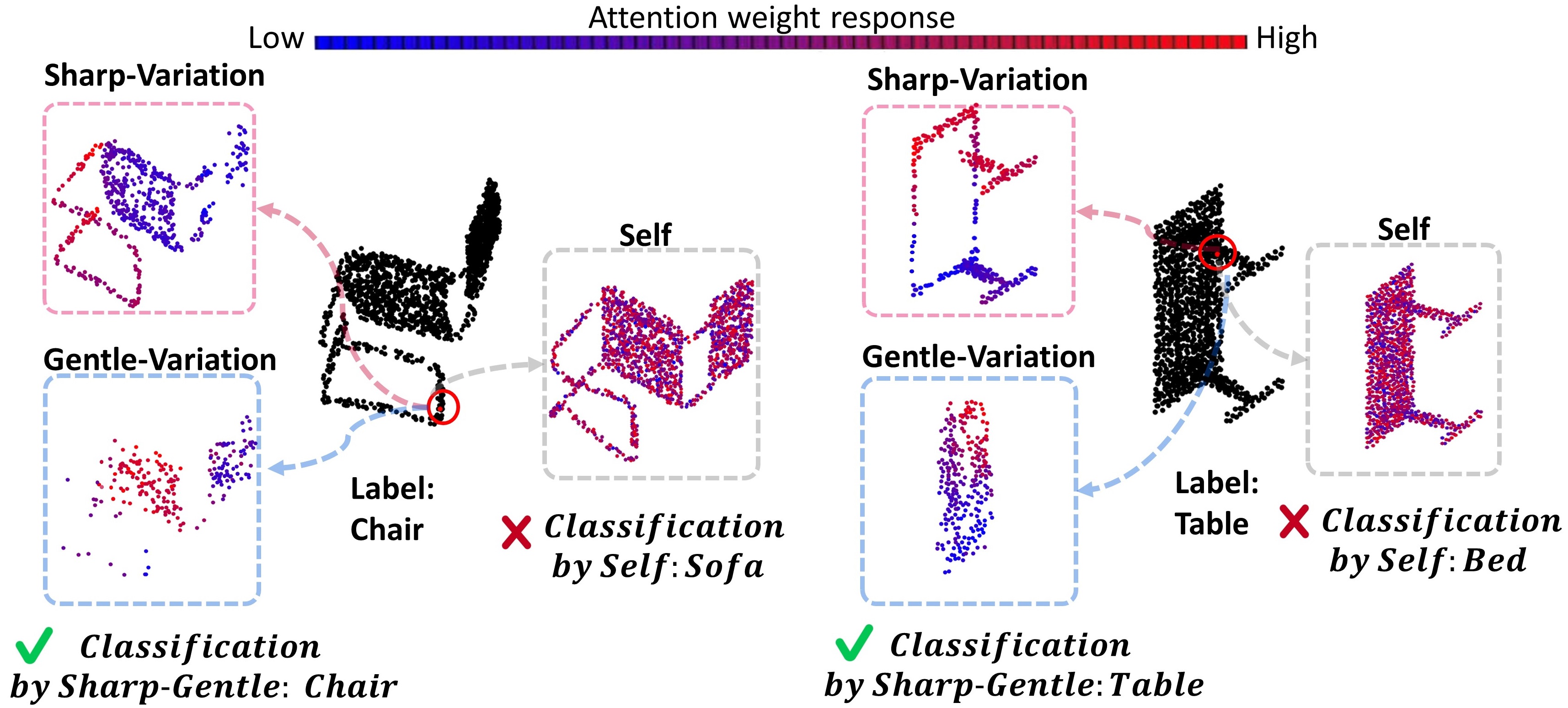}
	\end{center}
	\caption{Visualization of the attention weights distribution on Sharp-Gentle Complementary Attention and Self-attention. We select some points in original point clouds as the anchor points (in \textbf{circle}) to investigate their attention distributions, where the points drawn in \textbf{red} are assigned higher attention weights and the points drawn in \textbf{blue} are assigned lower attention weights with respect to anchor points. We observe that anchor points pay different attentions to points based on the geometric correlation.}
	\label{ATT}
\end{figure}
\subsection{4.3 Self-Attention or Sharp-Gentle Attention}
Self-Attention is an alternative holistic fusion strategy that pays different attentions to each point feature while linking them with the original point cloud itself, instead of integrating the feature from sharp and gentle variation components. 

However, self-attention unavoidably brings large redundancy due to it operates all points together, making it hard to capture the most related geometric interest to the network. Yet this issue is alleviated after disentangling point clouds, where our model easily pays different attentions to two variation components carried with few redundancy and distinct geometric information. 

To verify this, we compare our module with applying self-attention fusion on original point features and visualize the attention weights distribution in Fig.~\ref{ATT}. The attention weights in self-attention are assigned irregularly, which indicates that self-attention method is limited to capture the most relevant geometric information. By contrast, through our module, different point features from sharp and gentle variation components are assigned different weights based on the geometric correlation with the anchor point in the original input. As shown in Fig.~\ref{ATT}, a point in one leg of desk pays more attention to the points belongs to that leg among sharp-variation component and the relatively geometry-correlated points among gentle-variation component. 

Therefore, compared with self-attention, our attention module drives the network to easily capture the distinct and complementary geometric information with less redundancy from two disentangled components. The quantitative comparison is listed in Table \ref{ablation_study}.

\subsection{4.4 Geometry-Disentangled Attention Network}
As illustrated in Fig.~\ref{FAFNET}, we combine Geometry-Disentangle Module (GDM) and Sharp-Gentle Complementary Attention Module (SGCAM) as basic blocks to design Geometry-Disentangled  Attention Network for 3D point cloud analysis. Before each block, features of each point and its K-nearest neighbors (KNN) are concatenated through the local operator. In each block, an adjacency matrix is constructed in the feature space through GDM to disentangle points into sharp and gentle variation components. Then we fuse features from these two components with the original input of the block through SGCAM. A residual connection \cite{He_2016_CVPR} is applied at the end of each block. Two basic blocks are followed by another local operator. After the last MLP layer, the final global representation from max-pooling followed by fully connected and softmax layers is configured to the classification task. For the segmentation task, the outputs of the last MLP layer are directly passed through the fully connected as well as softmax layers to generate per-point scores for semantic labels. Note that each original input point is not only integrated with its nearest neighbors to capture local structure, but also linked with all of the disentangled sharp and gentle variation components that beyond the local area to describe distinct and complementary 3D geometries. Table \ref{ablation_study} shows the quantitative comparison of applying our modules with only using KNN.

%\noindent{\bf Dynamic adjacency matrix calculation.} 
\subsubsection{Dynamic Adjacency Matrix Calculation.}
Inspired by \cite{Wang:2019:DGC:3341165.3326362}, the adjacency matrix at the beginning of the GDM in each block is calculated in a dynamic way depending on the features learned in different semantic levels. Fig.~\ref{selection} (b) shows that our module successfully disentangles points into sharp and gentle components in various objects. This disentanglement module is jointly optimized during learning to help the network better model the geometric structure of objects. Table~\ref{ablation_study_3} in Sec 5.2 suggests the quantitative comparison of this dynamic operation with the operation of pre-selecting points before training.

\section{Experiments}

We evaluate our network on shape classification task and part segmentation task on various datasets. Furthermore, we provide other experiments to analyze our network in depth.

\subsection{5.1 3D Point Cloud Processing}\label{5.1} 
%\noindent{\bf Object classification.} 
\subsubsection{Object Classification.}
We firstly evaluate GDANet on ModelNet40 \cite{Wu_2015_CVPR}.~It contains 9843 training models and 2468 test models in 40 categories and the data is uniformly sampled from the mesh models by \cite{Qi_2017_CVPR}. Same with \cite{Wang:2019:DGC:3341165.3326362}, the training data is augmented by randomly translating objects and shuffling the position of points. Similar to \cite{Qi_2017_CVPR,NIPS2017_7095}, we perform several voting tests with random scaling and average the predictions during testing. Table \ref{Result_classification_M40} lists the quantitative comparisons with the state-of-the-art methods. GDANet outperforms other methods using only 1024 points as the input.

\begin{table*}[t]
	\begin{center}
			\resizebox{\textwidth}{!}{
			\begin{tabular} {l|c|c|cccccccccccccccc}
				\hline
				method (time order)  & class & inst. & aero & bag & cap & car & chair & ear & guitar & {  knife} & lamp & lap & motor & mug & pistol & rocket & { skate} & table  \\
				 &mIOU& mIOU  &   &   &   &   &   & phone &   &   &   & top   &   &   &  &   & board & \\
				\hline
				Kd-Net\cite{Klokov_2017_ICCV}  & 77.4 & 82.3 & 80.1 & 74.6 & 74.3 & 70.3 & 88.6 & 73.5 & 90.2 & 87.2 & 81.0 & 94.9 & 57.4 & 86.7 & 78.1 & 51.8 & 69.9 & 80.3\\
				PointNet\cite{Qi_2017_CVPR}   & 80.4 & 83.7 & 83.4 & 78.7 & 82.5 & 74.9 & 89.6 & 73.0 & {91.5} & 85.9 & 80.8 & 95.3 &  65.2 & 93.0 & 81.2 & 57.9 & 72.8 & 80.6 \\
				PointNet++\cite{NIPS2017_7095}& 81.9 & 85.1 & 82.4 & 79.0 & 87.7 & 77.3 & 90.8 & 71.8 & 91.0 & 85.9 & {83.7} & 95.3 & {71.6} & 94.1 & 81.3 & 58.7 & 76.4 & 82.6\\
				SyncCNN\cite{synccnn}  & 82.0 & 84.7 & 81.6 & 81.7 & 81.9 & 75.2 & 90.2 & 74.9 & \textbf{93.0} & 86.1 & 84.7 & 95.6 & 66.7 & 92.7 & 81.6 & 60.6 & 82.9 & 82.1\\
				SCN\cite{Xie_2018_CVPRSCN}  & 81.8 & 84.6  & 83.8 & 80.8 & 83.5 & 79.3 & 90.5 & 69.8 & 91.7 & 86.5 & 82.9 & 96.0 & {69.2} & 93.8 & 82.5 & 62.9 & 74.4 & 80.8\\
				KCNet\cite{Shen_2018_CVPRkc} & 82.2 &84.7  & 82.8 & 81.5 & 86.4 & 77.6 & 90.3 & 76.8 & 91.0 & 87.0 & 84.5 & 95.5 & 69.2 & 94.4 & 81.6 & 60.1 & 75.2 & 81.3\\
				SpiderCNN\cite{Xu_2018_ECCV}  & 82.4 & 85.3 & 83.5 & 81.0 & 87.2 & 77.5 & 90.7 & 76.8 & 91.1 & 87.3 & 83.3 & 95.8 & 70.2 & 93.5 & 82.7 & 59.7 & 75.8 & 82.8\\
				DGCNN\cite{Wang:2019:DGC:3341165.3326362}  & 82.3 & 85.2 & 84.0 & 83.4 & 86.7 & 77.8 & 90.6 & 74.7 & {91.2} & 87.5 & 82.8 & 95.7 & 66.3 & 94.9 & 81.1 & 63.5 & 74.5 & 82.6 \\
				RS-CNN\cite{liu2019rscnn} & 84.0 & 86.2 & 83.5 & 84.8 & 88.8 & 79.6 & \textbf{91.2} & 81.1 & 91.6 & 88.4 & \textbf{86.0} & 96.0 & 73.7 & 94.1 & 83.4 & 60.5 & 77.7 & 83.6\\
				DensePoint\cite{liu2019densepoint} & 84.2 & 86.4 & 84.0 & 85.4 & 90.0 & 79.2 & 91.1 & 81.6 & 91.5 & 87.5 & 84.7 & 95.9 & 74.3 & 94.6 & 82.9 & \textbf{64.6} & 76.8 & 83.7\\
				InterpCNN\cite{interconv} & 84.0 & 86.3 & - & - & - & - & - & - & - & - & - & - & - & - & - & - & - & -\\
				3D-GCN\cite{Lin_2020_CVPR} & 82.1 & 85.1 & 83.1 & 84.0 & 86.6 & 77.5 & 90.3 & 74.1 & 90.0 & 86.4 & 83.8 & 95.6 & 66.8 & 94.8 & 81.3 & 59.6 & 75.7 & 82.8\\
				GSNet(Xu et al. 2020) & 83.5 & 85.3 & 82.9 & 84.3 & 88.6 & 78.4 & 89.7 & 78.3 & 91.7 & 86.7 & 81.2 & 95.6 & 72.8 & 94.7 & 83.1 & 62.3 & 81.5 & 83.8\\	
				\textbf{GDANet(ours)}  & \textbf{85.0} & \textbf{86.5} & \textbf{84.2} & \textbf{88.0} & \textbf{90.6} & \textbf{80.2} & 90.7 & \textbf{82.0} & 91.9 & \textbf{88.5} & 82.7 &  \textbf{96.1} & \textbf{75.8}  &  \textbf{95.7} &  \textbf{83.9} & 62.9 & \textbf{83.1}  & \textbf{84.4}\\
				\hline
			\end{tabular}
			}
			\caption{Segmentation results (\%) on ShapeNet Part dataset.}
	\label{Result_sgm}
\end{center}
%\vspace{-0.1cm}
\end{table*}

\begin{table}
	\begin{center}
	\resizebox{0.46\textwidth}{!}{
	\begin{tabular}{lcc}
		\hline
		Method (time order) & Input & Acc.\\
		\hline
		PointNet\cite{Qi_2017_CVPR} & 1K points & 89.2\\
		PointNet++\cite{NIPS2017_7095} & 1K points  &  90.7\\
		SCN\cite{Xie_2018_CVPRSCN} & 1K points  & 90.0 \\
		KCNet\cite{Shen_2018_CVPRkc} & 1K points  & 91.0  \\
		PointCNN\cite{NIPS2018_7362} & 1K points  & 92.2 \\
		PointWeb\cite{pointweb} & 1K points  & 92.3 \\
		Point2Sequence\cite{point2sequence} & 1K points  & 92.6 \\
		DGCNN\cite{Wang:2019:DGC:3341165.3326362} & 1K points & 92.9 \\
		KPConv\cite{thomas2019KPConv} & 1K points & 92.9\\
		InterpCNN\cite{interconv} & 1K points & 93.0\\
		DensePoint\cite{liu2019densepoint}  & 1K points & 93.2\\
		Geo-CNN\cite{Lan_2019_CVPR} & 1K points & 93.4\\
		RS-CNN\cite{liu2019rscnn} & 1K points & 93.6\\
		3D-GCN\cite{Lin_2020_CVPR} & 1K points & 92.1\\
		FPConv\cite{fpconv} & 1K points  & 92.5 \\
		GSNet\cite{xu2020geometry} & 1K points  & 92.9 \\
		\textbf{GDANet(ours)}  & \textbf{1K points} & \textbf{93.8}\\
		\hline
		PointNet++\cite{NIPS2017_7095} & 5K points+normal& 91.9\\
		PointConv\cite{PointConv} & 1K points+normal & 92.5\\
		\hline
	\end{tabular}}
	\caption{Classification accuracy (\%) on ModelNet40 dataset.}
	\label{Result_classification_M40}
	\end{center}
\end{table}

\begin{figure}[t]
\setlength{\abovecaptionskip}{0cm} 
	\begin{center}
    \includegraphics[height=2cm]{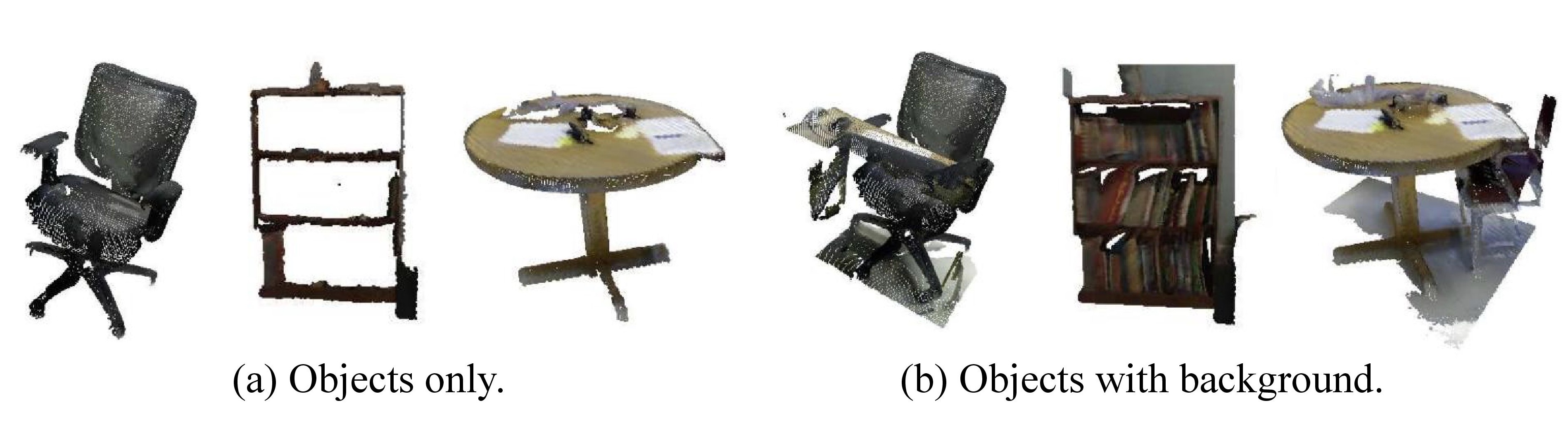}
    \end{center}
	\caption{Visualization of objects in ScanObjectNN.}
	\label{scanobj}
\end{figure}

Our model is also tested on ScanObjectNN by \cite{Uy_2019_ICCVnew}, which is used to investigate the robustness to \textbf{noisy} objects with deformed geometric shape and non-uniform surface density in the real world. We adopt our model on the OBJ$\_$ONLY (simplest variant of the dataset) and OBJ$\_$BG (more noisy background). Sample objects of these variants are shown in Fig.~\ref{scanobj}. We retrain the methods listed in \cite{Uy_2019_ICCVnew} and compare them with our network. The results are summarized in Table \ref{Result_classification_ScanOBJ}, where our model gets the highest accuracy and the lowest performance drop from OBJ$\_$ONLY to OBJ$\_$BG. This proves the practicality of our method in the real world.

\begin{table}[t]
	\centering
    \resizebox{0.46\textwidth}{!}{
	\begin{tabular}{p{130pt}ccccc}
		\hline
		Method & OBJ\_ONLY & OBJ\_BG&acc drop\\
		\hline
		PointNet\cite{Qi_2017_CVPR} & 79.2 & 73.3& $\downarrow 5.9$\\
		SpiderCNN\cite{Xu_2018_ECCV} & 79.5 & 77.1&$\downarrow 5.4$\\
		PointNet++\cite{NIPS2017_7095} & 84.3 & 82.3&$\downarrow 2.0$\\
		DGCNN\cite{Wang:2019:DGC:3341165.3326362} & 86.2 & 82.8&$\downarrow 3.4$\\
		PointCNN\cite{NIPS2018_7362} & 87.9 & 85.8 &$\downarrow 2.1$\\
		\textbf{GDANet(ours)} & \textbf{88.5} & \textbf{87.0} &$\downarrow 1.5$\\
	\hline
\end{tabular}}
	\caption{Classification results (\%) on ScanObjectNN dataset (noise robustness test).}
    \label{Result_classification_ScanOBJ}
\end{table}
%\end{table}

%\noindent\textbf{Shape Part Segmentation.}
\subsubsection{Shape Part Segmentation.}
\begin{figure}[t]
\setlength{\abovecaptionskip}{0cm} 
	\begin{center}
	\includegraphics[height=1.3cm]{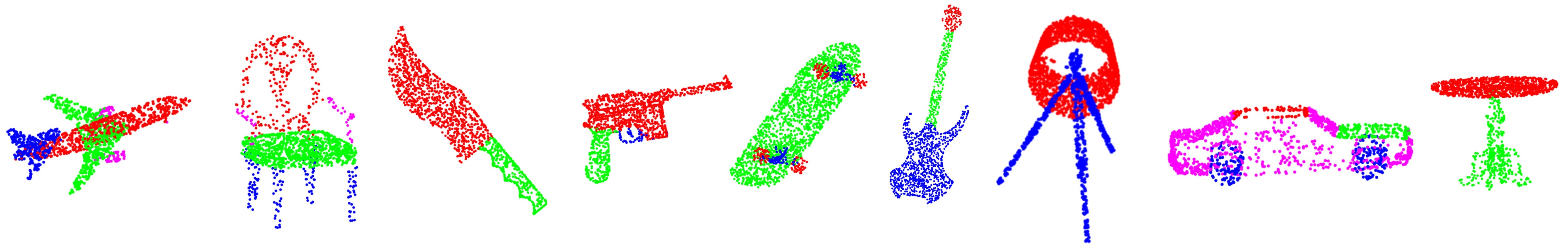} 
	\end{center}
	\caption{Part Semantic Segmentation examples.}
	\label{seg_vis}
\end{figure}

Shape Part segmentation is a more challenging task for fine-grained shape recognition. We employ ShapeNet Part \cite{Yi:2016:SAF:2980179.2980238} that contains 16881 shapes with 16 categories and is labeled in 50 parts where each shape has 2\-5 parts. Our network is trained with multiple heads to segment the parts of each object categories. Same voting test in classification is conducted. Table \ref{Result_sgm} summarizes the instance average, the class average and each class mean Inter-over-Union (mIoU). GDANet achieves the best performance with class mIoU of 85.0$\%$ and instance mIoU of 86.5$\%$. It is worth mentioning that GDANet performs better on objects with obvious geometric structure such as bag, mug and table. Fig.~\ref{seg_vis} visualizes some segmentation results.

\begin{figure}[t]
	\begin{center}
	\includegraphics[width=0.85\columnwidth]{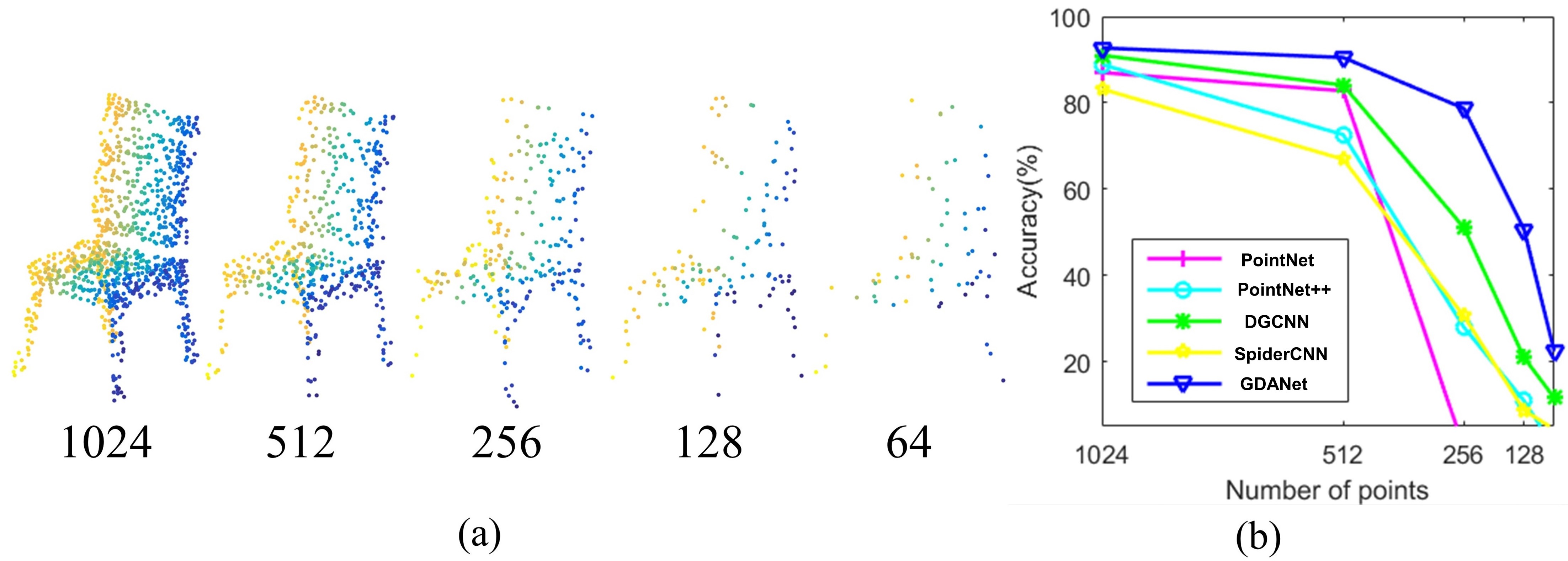}
	\end{center}
	\caption{~Density robustness test. (a). Point cloud with random point dropout. (b). Test results on ModelNet40 of using sparser points as the input to a model
trained with 1024 points. For fair comparisons, all methods have no data enhancement of random point dropout during training.}
	\label{droupt}

\end{figure}
\begin{table}[t]
\begin{center}
\resizebox{0.3\textwidth}{!}{
\begin{tabular}{cccccc}
		\hline
		knn & self & sharp & gentle &voting & acc. (\%)\\
		\hline
		&&&&&91.5\\
		\checkmark&&&&&92.2\\
		&\checkmark&&&&91.7\\
		\checkmark&\checkmark&&&&92.6\\
		&&\checkmark &  \checkmark & & 92.7 \\
		\checkmark& &\checkmark & &   &  92.7 \\
		\checkmark&&   & \checkmark & &  92.4 \\
	    \checkmark&& \checkmark & \checkmark && \textbf{93.4} \\
		\checkmark&& \checkmark & \checkmark &\checkmark& \textbf{93.8}\\
		\hline
	\end{tabular}}
	 \caption{Geometry-Disentangled complementary effect to supplement KNN information in GDANet on ModelNet40. ~`knn' indicates KNN aggregation, ~`self' means the input point cloud is fused with itself by self-attention, ~`sharp' and ~`gentle' denote the input point cloud is fused with features of sharp and gentle variation, ~`voting' is the voting strategy during testing, respectively.}
	\label{ablation_study}
	\end{center}
\end{table}

\begin{table}[t]
\begin{center}
\resizebox{0.35\textwidth}{!}{
	\begin{tabular}{lcccc}
		\hline
		method & ModelNet40 & OBJ\_ONLY & OBJ\_BG\\
		\hline
		random   &  92.6 & 84.7 & 84.3\\
		FPS &  92.7 & 86.0 & 84.3\\
		\textbf{sharp-gentle}  &  \textbf{93.8} & \textbf{88.1}  & \textbf{86.6}\\
		\hline
	\end{tabular}}
	\caption{Classification results (\%) of using different point selection methods in our Geometry-Disentangle Module.}
	\label{ablation_study_2}
\end{center}
\end{table}

\subsection{5.2 Network Analysis}\label{5.2} 
%\noindent{\bf Ablation Study.} 
\subsubsection{Ablation Study.}
The results are summarized in Table \ref{ablation_study}. When the input point cloud is fused with features from both sharp and gentle variation components, the network achieves the best accuracy with 93.4$\%$. GDANet also surpasses the architecture of only using KNN with 1.2$\%$\(\uparrow\). Eventually, our method obtains an accuracy of 93.8$\%$ with voting tests. This experiment supports our claim that the disentangled sharp and gentle variation components cooperatively provide different and complementary geometric information to supplement KNN local semantics.
Furthermore, we replace the selection of sharp and gentle variation components with random and Furthest Point Selection (FPS) in GDM. The results are listed in Table~\ref{ablation_study_2}, where disentangling point clouds into sharp and gentle variation components gets the highest accuracy and is noticeably robust on the noisy dataset ScanObjectNN \cite{Uy_2019_ICCVnew}. Empirically, our disentanglement strategy selects points carried with informative geometries instead of noisy points, which improves the noise robustness. 
\begin{table}[t]
\begin{center}	
	\resizebox{0.2\textwidth}{!}{
\begin{tabular}{ccccc}
		\hline
		method & acc. (\%)\\
		\hline
		 Precomputed  & 93.0 \\
		 Dynamic & \textbf{93.8} \\
		\hline
	\end{tabular}}
	\caption{Impact of dynamic strategy.}
	\label{ablation_study_3}
	\end{center}

\begin{center}
	\resizebox{0.3\textwidth}{!}{
	\begin{tabular}{lccccc}
		\hline
		number & 1024 & 512 & \textbf{256} & 128\\
		\hline
		acc. (\%) & 92.6 & 93.2 & \textbf{93.8} & 92.9\\
		\hline
	\end{tabular}}
		\caption{Selecting different number of points in GDM on ModelNet40.}
	\label{M_FF}
\end{center}

\begin{center}
\resizebox{0.3\textwidth}{!}{
	\begin{tabular}{cccc}
		\hline
		Method  & z/z & s/s\\
		\hline
		PointNet\cite{Qi_2017_CVPR} &  81.6 & 66.3\\
		PointNet++\cite{NIPS2017_7095} &  90.1 & 87.8\\
		SpiderCNN\cite{Xu_2018_ECCV}  &  83.5 & 69.6\\
		DGCNN\cite{Wang:2019:DGC:3341165.3326362} &  90.4 & 82.6\\
		\hline
		\textbf{GDANet(ours)} & \textbf{91.2} & \textbf{90.5}\\
		\hline
	\end{tabular}}
	\caption{Accuracy (\%) comparisons of rotation on ModelNet40. z/z: both training and test sets are augmented by random rotation for z axis; s/s: both training and test sets are augmented by random rotation for three axis (x,y,z).}
	\label{Result_rotation}
	\end{center}
\begin{center}
\resizebox{0.4\textwidth}{!}{
	\begin{tabular}{p{130pt}cc}
		\hline
		\text { Method } & {\#\text {params }} & {acc. (\%)} \\
		\hline
		\text { PointNet\cite{Qi_2017_CVPR} }  & {3.50 {M}} & {89.2} \\
		{\text { PointNet++\cite{NIPS2017_7095} }}  & {1.48 {M}} & {90.7} \\
		{\text { KPConv\cite{thomas2019KPConv}}}  & {6.15 {M}} & {92.9}\\
	
		{\text { DGCNN\cite{Wang:2019:DGC:3341165.3326362} }} &{1.81 {M}} & {92.9} \\
		{\text { GSNet\cite{xu2020geometry}}} & {1.51 {M}} & {92.9} \\
		{\textbf { GDANet(ours)}} & \textbf{0.93 {M}} & \textbf{93.8}  \\
		\hline
	\end{tabular}}
	\caption{Comparisons of model complexity on ModelNet40.}
	\label{Result_Complexity}
\end{center}
\end{table}

Moreover, the result of dynamic adjacency matrix calculation (Sec 4.4) is shown in Table~\ref{ablation_study_3}. We pre-compute the adjacency matrix on 3D coordinates to pre-disentangle different variation components, and fuse the features from them, which gets 93.0\% accuracy. Yet by dynamically calculating the adjacency matrix on semantic features in GDM at different levels, our network gains 0.8$\%$\(\uparrow\). Our disentanglement module is jointly optimized during training for modeling different geometric structures. Fig.~\ref{selection} shows that our disentanglement strategy successfully decompose points into sharp (contour) and gentle (flat area) variation components. 

Last, we investigate the impact of the number of selected points M in GDM, which is shown in Table \ref{M_FF}. GDANet performs best when selecting 25\% of input points, which proves the benefit of our disentanglement strategy. When the number of selected points equals to the number of input, it indicates self-attention. % More investigations of GDANet are included in the supplementary material.
% \begin{table}
% \begin{center}
% \tiny
% 	\begin{tabular}{lccccc}
% 		\hline
% 		\# selected points & 1024 & 512 & \textbf{256} & 128\\
% 		\hline
% 		acc. (\%) & 92.6 & 93.8 & \textbf{93.8} & 92.9\\
% 		\hline
% 	\end{tabular}
% 		\caption{GDANet under different number of selected points on ModelNet40.}
% 	\label{M_FF}
% 		\end{center}
% \end{table}

%\noindent{\bf Robustness Analysis.} 
\subsubsection{Robustness Analysis.}
First, the robustness of our network on sampling density is tested by using sparser points as the input to GDANet trained with 1024 points.
Fig.~\ref{droupt} shows the result. Although sparser points bring more difficulties, GDANet still performs consistently at all density settings.

Moreover, Table \ref{Result_rotation} summarizes the results of rotation robustness, where GDANet performs best especially with 2.7$\%$\(\uparrow\) than the second best at (s/s).

% \begin{table}[t]
% \begin{center}
% \tiny
% 	\begin{tabular}{cccc}
	
% 		\hline
% 		Method  & z/z & s/s\\
% 		\hline
% 		PointNet\cite{Qi_2017_CVPR} &  81.6 & 66.3\\
% 		PointNet++\cite{NIPS2017_7095} &  90.1 & 87.8\\
% 		SpiderCNN\cite{Xu_2018_ECCV}  &  83.5 & 69.6\\
% 		DGCNN\cite{Wang:2019:DGC:3341165.3326362} &  90.4 & 82.6\\
% 		\hline
% 		\textbf{GDANet(ours)} & \textbf{91.2} & \textbf{90.5}\\
% 		\hline
% 	\end{tabular}
% 	\caption{Accuracy (\%) comparisons of rotation on ModelNet40. z/z: both training and test sets are augmented by random rotation for z axis; s/s: both training and test sets are augmented by random rotation for three axis (x,y,z).}
% 	\label{Result_rotation}
% \end{center}
% \end{table}

% \begin{table}[t]

% \end{table}

Last, our model is tested on ScanObjectNN \cite{Uy_2019_ICCVnew} that consists of noisy objects with deformed geometric shape and non-uniform surface density. Table \ref{Result_classification_ScanOBJ} shows GDANet gains the lowest accuracy drop from OBJ\_ONLY to OBJ\_BG, which proves the noise robustness of GDANet. 

%\noindent{\bf Model Complexity.} 
\subsubsection{Model Complexity.}
Table~\ref{Result_Complexity} shows the complexity by comparing the number of parameters. GDANet reduces the number of parameters by 84.9$\%$ and increases the accuracy with 0.9$\%$\(\uparrow\) compared with KPConv \cite{thomas2019KPConv}. 

\section{Conclusion}
This work proposes GDANet for point cloud processing. Equipped with Geometry-Disentangle Module, GDANet dynamically disentangles point clouds into sharp and gentle variation components in different semantic levels, which respectively denotes the contour and flat area of a point cloud. Another core component is Sharp-Gentle Complementary Attention Module, which applies the attention mechanism to explore the relations between original points and different variation components to provide geometric complementary information for understanding point clouds. Extensive experiments have shown that our method achieves state-of-the-art performances and decent robustness.

\section{Acknowledgments}
This work was supported in part by Guangdong Special Support Program under Grant (2016TX03X276), in part by the National Natural Science Foundation of China under Grant (61876176, U1713208), and in part by the Shenzhen Basic Research Program (CXB201104220032A), the Joint Laboratory of CAS-HK. This work was done during Mutian Xu's internship at Shenzhen Institutes of Advanced Technology, Chinese Academy of Sciences.

%\clearpage

%{\small
%\bibliographystyle{aaai}
%\bibliography{bibFile}
%}

%\end{document}

% \begin{quote}
% \begin{small}
% \bibliographystyle{aaai21}
% \bibliography{bibfile}
% \end{small}
% \end{quote}

% \end{document}

{\small

%\bibliography{Formatting-Instructions-LaTeX-2021}
}
\end{document}